\documentclass[11pt,letter]{article}
\usepackage[hyperref]{eacl2021}
\usepackage{times}
\usepackage{latexsym}

\usepackage{microtype}

\def\aclpaperid{NN} %  Enter the acl Paper ID here
\aclfinalcopy

\usepackage{url}
\usepackage{booktabs}       % professional-quality tables
\usepackage{amsfonts}       % blackboard math symbols
\usepackage{nicefrac}       % compact symbols for 1/2, etc.

\usepackage{eqnarray,amsbsy,amsmath,amssymb,amsthm}
\usepackage{graphicx}
\usepackage{algorithm2e}
\usepackage{multirow}

\usepackage{bbm}

\usepackage{csquotes}
\usepackage{mathtools}
\usepackage{comment}
\usepackage{natbib}
\usepackage{enumitem}
\usepackage{xspace}

\newcommand{\qmds}{\textsc{AQuaMuSe}\xspace}
\newcommand{\nq}{\textsc{NQ}\xspace}
\newcommand{\cc}{\textsc{CC}\xspace}

\newcommand{\old}[1]{}
\newcommand{\SC}[1]{{\color{cyan}SC: #1}}

\newcommand{\highlightI}[1]{{\color{blue}#1}}
\newcommand{\highlightII}[1]{{\color{magenta}#1}}
\newcommand{\hiddenHighlight}[1]{{\color{gray}#1}}

\newcommand{\eat}[1]{}

\usepackage{soul}
\usepackage{color}
\definecolor{aquamarine}{rgb}{0.5, 1.0, 0.83}
\definecolor{Mycolor2}{HTML}{00F9DE}

\usepackage{multirow}
\usepackage{tikz}
\usetikzlibrary{positioning,chains,fit,shapes,calc}
\usepackage{pgfplots}

\definecolor{turquoise}{cmyk}{.43,0,.24,0} 
\definecolor{myblue}{RGB}{80,80,160}
\definecolor{mygreen}{RGB}{143,255,204}

\tikzstyle{plate caption} = [caption, node distance=0, inner sep=-0.pt,below left=1pt and 0pt of #1.south east]
\tikzstyle{plate} = [draw, rectangle, fit=#1]

\graphicspath{{pictures/}}

\usetikzlibrary{calc}
\usepackage{zref-savepos}
\usepackage{tabu}

\title{\qmds: Automatically Generating Datasets for  Query-Based Multi-Document Summarization}

\author{Sayali Kulkarni \\
Google Research\\
   \texttt{sayali@google.com} \\\And
 Sheide Chammas \\
 Google Research\\
   \texttt{sheide@google.com} \\\And
 Wan Zhu \\
 Google Research\\
   \texttt{wanzhu@google.com } \\\AND
 Fei Sha  \\
 Google Research\\
   \texttt{fsha@google.com} \\\And
 Eugene Ie  \\
 Google Research\\
   \texttt{eugeneie@google.com} \\
}

\date{}

\pgfplotsset{compat=1.15}
\begin{document}

\maketitle
\begin{abstract}
Summarization is the task of compressing source document(s) into coherent and succinct passages. This is a valuable tool to present users with concise and accurate sketch of the top ranked documents related to their queries. Query-based multi-document summarization (qMDS) addresses this pervasive need, but the research is severely limited due to lack of training and evaluation datasets as existing single-document and multi-document summarization datasets are inadequate in form and scale. We propose a scalable approach called \qmds to automatically mine qMDS examples from question answering datasets and large document corpora. Our approach is unique in the sense that it can general a dual dataset --- for extractive and abstractive summaries both. We publicly release a specific instance of an \qmds dataset with 5,519 query-based summaries, each associated with
an average of 6 input documents selected from an index of 355M documents from Common Crawl\footnote{\url{https://commoncrawl.org}}. Extensive evaluation of the dataset along with baseline summarization model experiments are provided.

\end{abstract}

% !TEX root = main.tex

\section{Introduction}
\label{sIntro}

Summarization has been a challenging problem in natural language processing. Recently a number of neural encoder-decoder approaches have made significant progress in this research area \cite{rush2015neural, see2017get, wang2018reinforced}.  State-of-the-art models such as PEGASUS \cite{Zhang2019PEGASUSPW} have leveraged related data sources and tasks, such as language modeling, to pre-train massive summarization models. But only a few large-scale high-quality human-curated summarization datasets available for training and evaluation.

Obtaining human annotation for summarization is a nontrivial task in itself and there are several contributing factors. Summarization is often subjective and depends on the annotators' reading comprehension abilities (especially on unfamiliar topics), their interpretation of the text and their judgement on what piece of information should be considered important or relevant to the use case of the generated summaries. These are all influenced by the annotators' own life experiences, and in the case of abstractive summarization, their ability to compose fluent and succinct text passages as well. In some scenarios, such as generating news headlines or identifying how-to instructions (as a form of summary outline), it is possible to obtain and repurpose pre-annotated data from established web publishers for single document summarization (SDS) \cite{hermann2015teaching,koupaee2018wikihow}. However it's less clear how one would approach data collection for multi-document summarization (MDS) from the open web.

\begin{figure*}
  \centering
    \includegraphics[width=0.9\textwidth, viewport=120 225 970 525, clip=true]{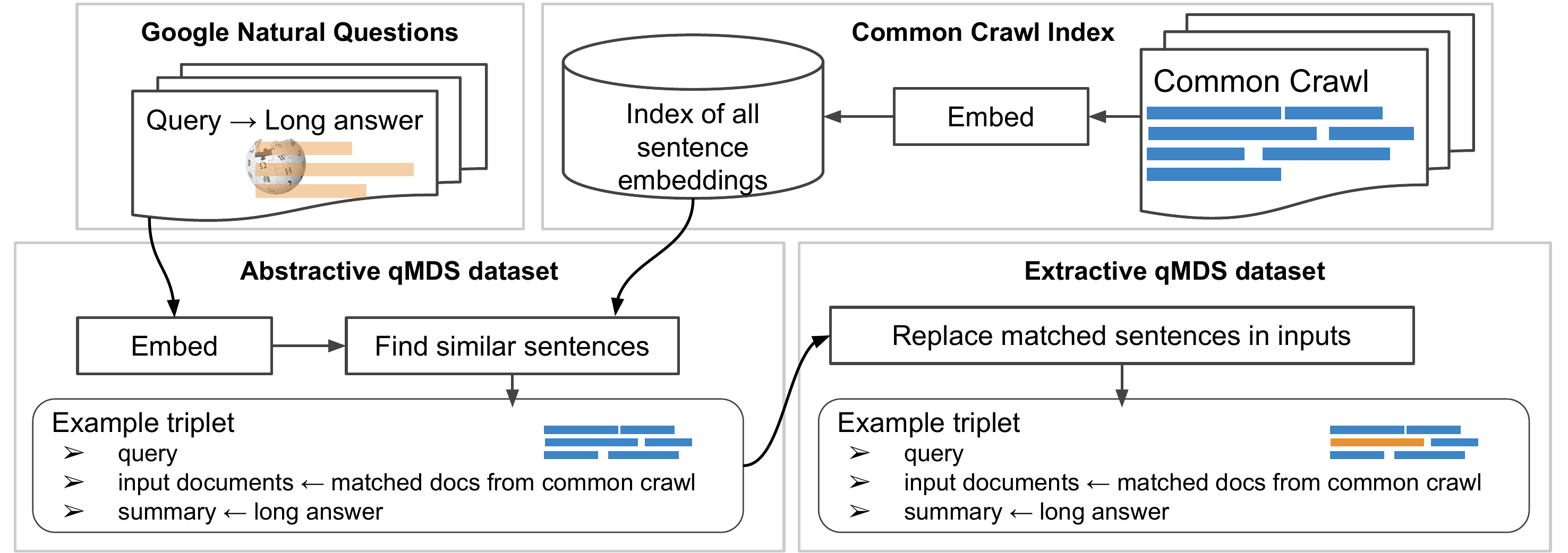}
  \caption{\qmds pipeline for generating conjugate abstractive and extractive query based multi-document summarization datasets.}
\label{fig:qmds_arch}
\end{figure*}

Two recent studies, WikiSum and Multi-News, have attempted to tackle this problem with automatic procedures to harvest documents for MDS by crawling hyperlinks from Wikipedia and \texttt{newser.com} web sites respectively \cite{liu2018generating,fabbri2019multi}. Both studies target significantly longer summaries than short texts or headlines. However, the area of generating focused summaries conditioned on contexts has not been in the limelight. This is an important problem in natural language generation, for example in personalized news feed summaries, context-driven product review summaries, to name a few. 

Our work considers this variant of MDS called query-based MDS (qMDS) which have crucial applications in augmenting information retrieval (IR) experiences \cite{daume-iii-marcu-2006-bayesian,litvak-2017-query,abstractive-nn-qbs,baumel2018}. Text documents are typically multi-faceted and users are often interested in identifying information that is most relevant to their stated preferences. For example, suppose a user is interested in car reliability and cars from certain manufacturers, then an effective IR system could consolidate car reviews from across the web and provide concise summaries relating to the reliability of those cars of interest. Similar to MDS, qMDS also suffers from the lack of large-scale annotated data, especially for generating long abstractive summaries \cite{nema-etal-2017-diversity}. While we mainly focus on qMDS, the proposed dataset generation methodology can be re-adapted for the more general problem of MDS (while the reverse is not necessarily true).

Our contributions in this paper are two-folds. We first introduce a general approach for machine generating query-based multi-document summarization (qMDS) datasets at scale, with knobs to control the automatically generated outputs along dimensions such as document diversity and degree of relevance of the target summary. Second, we provide an automatically generated large-scale dataset for qMDS that we validate with baseline summarization experiments and human rater evaluations.

As aforementioned, we focus on summarizing several documents as multi-faceted answers to complex queries.  To this end, we leverage the publicly released Google Natural Questions (NQ) dataset \cite{NQ}, which contains real user queries from Google search logs, capturing a wide range of topics that interest people. Many questions have short answers (e.g., one or more entity names, or dates) derived from Wikipedia pages and have been used to form \nq question answering dataset for training a SQuAD-like span-based QA system. More importantly, a sizable portion of the questions are paired with long-form answers (e.g., paragraphs) that are vetted by human raters. These long-form answers address user questions with content that are focused and coherent. As the Wikipedia passages have also been read and edited by the Wikipedia readership, the writing of the passages should be of adequate quality.

As we are primarily interested in the qMDS setting for general web IR applications, we would like to simulate how a search engine might synthesize documents of high relevance to a user query. To identify high quality passages from the web that can be used to recreate target Wikipedia paragraphs, we use a pre-processed and cleaned version of the Common Crawl corpus \cite{2019t5} as a proxy web search index to select documents relevant to the \nq long-form answers. We take special considerations in including documents of varying semantic relevance such that our baseline task involves deriving summaries from documents with enough distracting information to challenge summarization models. Furthermore we ensure the sources are sufficiently diverse among themselves, as our primary interest is to summarize multi-faceted information. Figure \ref{fig:qmds_arch} illustrates the overall data generation procedure. We publicly release an instance of such a dataset containing 5,519 qMDS examples, that we split into training, validation and tests sets of sizes 4,555, 440 and 524 respectively\footnote{\url{https://github.com/google-research-datasets/aquamuse}}. Each example contains an average of 6 source documents to be synthesized into a long-form answer.

It is worth noting that the approach used in \citet{liu2018generating} to harvest Wikipedia article texts as summary targets is related to ours but there are a few key distinctions in goal and methodology. In the foremost, we aim to provide high-quality dataset for the task of qMDS (not just MDS). As such, we aim to generate paragraphs that are more consistent and coherent than generating full length articles, which have much stronger variability in both structure and content. The use of cited references in the WikiSum dataset do not necessarily provide adequate coverage for the summaries especially if some sentences in the Wikipedia text are missing references. Instead we use crawled documents from the web as potential source material. Since these are ``naturally occurring'' documents from the web (albeit a cached subset), we are simulating a realistic web IR application scenario across a large document corpus.

% !TEX root = main.tex
\section{\qmds}
\label{sApproach}
 The qMDS problem is formalized as follows. Given a query $q$, a set of related documents $R=\{r_i\}$, document passages $\{r_{i,j}\}$ relevant to the query are synthesized to an answer $a$. Various MDS approaches synthesize $a$ such that it is succinct and fluent natural language text that covers the information content in $\{r_{i,j}\}$ rather than just a concatenation of relevant spans. Such synthesized answers can augment information retrieval (IR) applications by enhancing the user experience with high-level query specific summaries.

We propose an automated approach to generating large datasets for the qMDS task for training and evaluating both abstractive and extractive approaches. We illustrate our approach using Google's Natural Questions (\nq) and Common Crawl (\cc). But the methodology is general enough to be extended to any other question answering dataset (containing answers that span multiple sentences) and web corpora (to serve as the domain for retrieval).

Google's \nq is an open-domain question answering dataset containing 307,373 training examples, 7,830 development examples, and 7,842 test examples in  version 1.0\cite{NQ}. Each example is a Google search query (generated by real users) paired with a crowd sourced short answer (one or more entities) and/or long answer span (typically a paragraph) from a Wikipedia page. Queries annotated with \emph{only} a long answer span serve as summarization targets since these cannot be addressed tersely by entity names (e.g. \textit{``Who lives in the Imperial Palace in Tokyo?''}) or a boolean. These queries result in open-ended and complex topics answers (e.g., \textit{``What does the word China mean in Chinese?''}).

\subsection{Approach}
Suppose a long answer is comprised of $n$ sentences $a=[l_1,...,l_n]$ and a document corpus $D=\{d_i\}$ consists of sentences $[d_{i,j}]$. We use the Universal Sentence Encoder \cite{universal-encoder} ($\phi$) to encode sentences $\phi(l_k)$ and $\phi(d_{i,j})$ for semantic similarity comparisons (e.g., using a dot product $s_{k,i,j} = \langle\phi(l_k), \phi(d_{i,j})\rangle$). This yield in partial result sets $R_k = \{(d_i, s_{k,i,j}): \theta_U > s_{k,i,j} > \theta_L \}$. These sets $R_{1..n}$ are then combined by $\psi(d_i) = \sum_{k,j} s_{k,i,j}$ to yield document-level scores to get result set $R = \{(d_i, \psi(d_i))\}$. We restrict the result set by selecting the top-K ranked documents. While we have made specific choices for $\phi$, $s_{k,i,j}$, $\psi$, they can be customized and tuned to construct result sets $R$ with higher/lower diversity, tighter/looser topicality match, or number of documents retrieved (to name a few) for evaluating summarization approaches under different qMDS task conditions.

With appropriate tuning of $\theta_U$ and $\theta_L$, the process above admits documents with sentences of varying semantic relevance into the result set $R$. Lowering $\theta_U$ (while keeping $\theta_L$ high enough) ensures we don't retain sentences $d_{i,j}$ that are exact matches of $l_k$ (but of at least some semantic equivalence), thereby generating qMDS abstractive summarization examples $(q, a, R)$. The relationship between $q$ and $R$ is transitive through the annotated long answer span $a$. For constructing extractive qMDS examples, we perform an in-place substitution of $d_{i,j}$ with $l_k$ (that can optionally be sampled according to match score $s_{k,i,j}$ in future work).

\begin{table*}
\small
    \centering
    \begin{tabular}{r|r|rrr|rrrrrr}
 & & \multicolumn{3}{c|}{\# examples} & \multicolumn{2}{c}{summary} &  \multicolumn{2}{c}{inputs} & \multicolumn{2}{c}{per-input doc} \\
Dataset & \# queries & train & dev & test & \# words & \# sents & \# words & \# sents & \# words & \# sents \\
\hline
\qmds & 5,519 & 4,555 & 440 & 524 & 105.9 & 3.8 & 9,764.1 & 405.7 & 1,597.1 & 66.4 \\
Debatepedia & 13,719 & 12,000 & 719  & 1,000  & 11.2 & 1 & 75.1 & 5.1 & 75 & 5.1 \\
Multi-News & NA &  44,972 & 5,622 & 5,622 & 263.7 & 10.0 & 2,103.5 & 82.7 & 489.2 & 23.4  \\
CNN/DM & NA & 287,227 & 13,368 & 11,490 & 56.2 & 3.7 & 810.6 & 39.8 & NA & NA \\
    \end{tabular}
     \caption{Comparison of recently proposed summarization datasets.}
    \label{tab:compare_ds}
\end{table*}

\subsection{Implementation details}
We use a pre-processed and cleaned version of the English \cc corpus called the Colossal Clean Crawled Corpus \cite{2019t5}. It contains 355M web pages in total. For the question answering data source, we use a 62.5\% sample of the NQ dataset from the train and development splits, in which 8.2\% are question answering examples that we matched with the CC corpus. These NQ questions are marked ``good'' by a majority of \nq raters and are paired with long-form answers. We limit to question answering pairs that cannot be addressed by terse responses (e.g. factoids), to simulate realistic qMDS use cases.

Using TensorFlow Hub\footnote{\url{https://tfhub.dev/}} we compute Universal Sentence Embeddings (which are approximately normalized) for sentences tokenized from both \nq and \cc data sources. The encoded \cc sentences are around 11Tb on disk while the \nq portion that formulates the target summaries is comparatively negligible in size. An exhaustive all pairwise comparison is performed using Apache Beam\footnote{\url{https://beam.apache.org/}}. The sentences from the \nq long answers are matched with the \cc corpus using efficient nearest neighbor searches over sentence embeddings indexed by space partitioning trees \cite{Liu2004AnIO}.

\begin{figure}
  \centering
    \includegraphics[width=0.35\textwidth, viewport=20 20 355 350, clip=true]{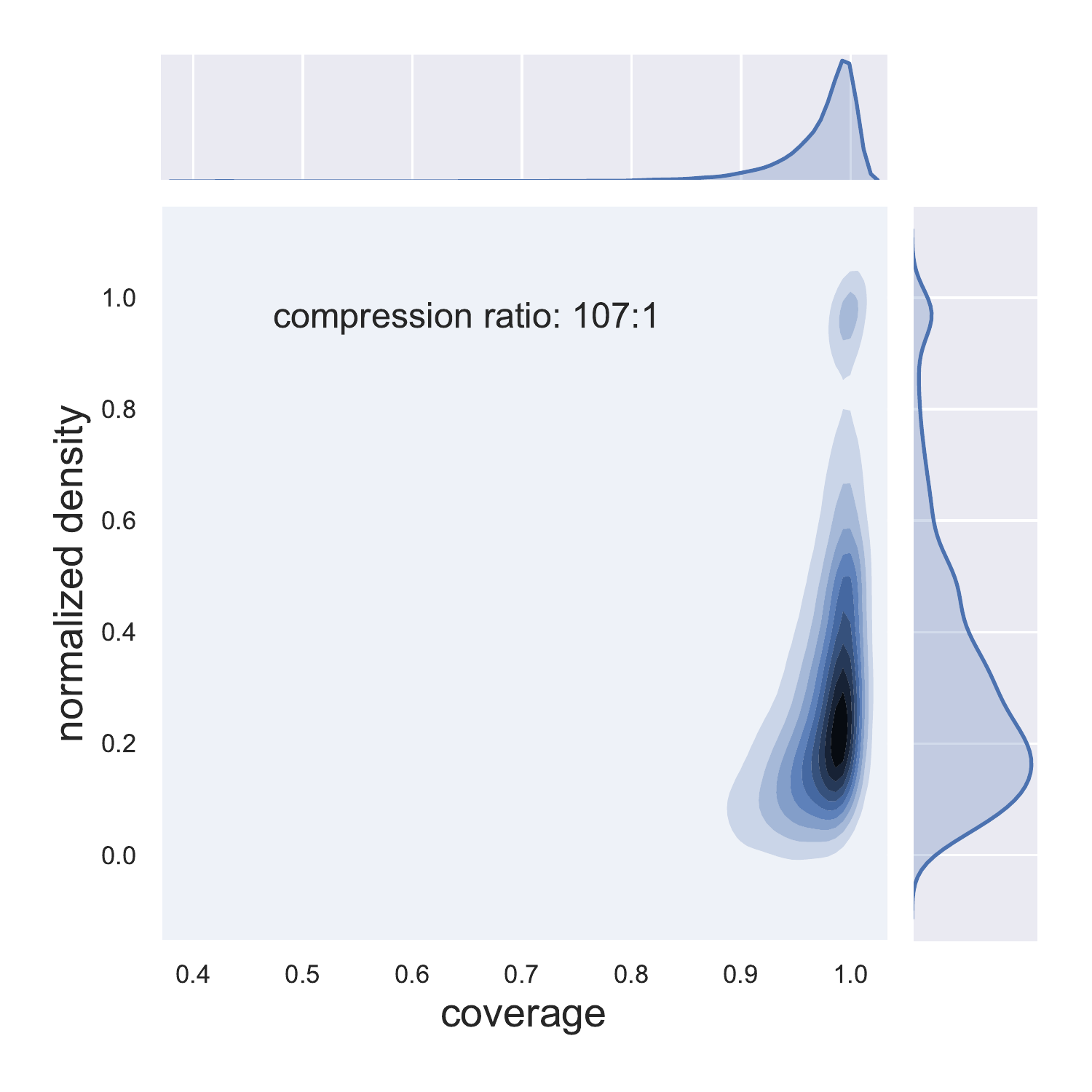}
    \caption{Coverage versus normalized density plot for the abstractive dataset shows the variance in the summary compared to inputs. The high compression ratio coming from long input documents would create a unique challenge for qMDS models.}
\label{fig:cd_plot}
\end{figure}

\paragraph{Sentence Matching Thresholds} $\theta_U$ and $\theta_L$ control the semantic relevance of sentences matched between \cc and \nq. Sentence pairs with matching scores below 0.8 are filtered out ($\theta_L$). To avoid exact sentence matches from pages with near-Wikipedia duplicates, we also filter out sentence pairs with scores above 0.99 ($\theta_U$). The \cc document match score is based on the sum of these sentence-to-sentence match scores. This can be used to trade-off the quality of the matched documents and the abstractive nature of the task. We use the coverage and density metrics defined in \citet{newsroom} to construct the normalized bivariate density plot illustrated in Figure \ref{fig:cd_plot}.

\paragraph{Summary Recall} It is entirely possible that we cannot locate a match for every sentence in a long answer. The \emph{Summary Recall} is the fraction of sentences in a long answer with matching \cc document sentences. A summary recall of 1.0 guarantees a summary can be generated. In our specific dataset instance, we restrict the summary recall to 0.75. Though this may seem like a handicap, we observed in experiments that the input documents have enough information to reconstruct the target summary fully and can be used to test the language generation capabilities of qMDS approaches.

\paragraph{Top-K Parameter} This is analogous to the number of top results returned by web search and controls the degree of support as well as diversity across multiple documents for a qMDS task. We evaluated the quality of the matched documents as ranked by their match scores. We use $K=7$ as we found document quality degrade after that (given our specific settings of sentence matching threshold and summary recall).

\subsection{Dataset Statistics}
\label{sApproachDatasetStatistics}
Our specific dataset instance is derived from a subset of the \nq dataset that we match with \cc. Based on the thresholds detailed above, we construct 5,519 examples that are split into training, development and testing sets (4,555, 440 and 524 examples, respectively) using a hash of the \nq long answer. Given the thresholds $\theta$ and $K$, the total number of \cc documents that matched our restricted set of \nq long answers is 33,760. Not every example can find up to $K=7$ matching documents from \cc within the $\theta$ bounds. The distribution of input document count are as shown in Table \ref{tab:megadoc_distri}.

\eat{\qmds generates both abstractive and extractive datasets for qMDS.}

\begingroup
\setlength{\tabcolsep}{4pt}
\begin{table}
\small
    \centering
    \begin{tabular}{c|ccccccc}
\small{\# input docs} & 1 & 2 & 3 & 4 & 5 & 6 & 7 \\
\hline
\small{\# queries} & 198 & 248 & 270 & 258 & 221 & 229 & 4,095 \\
    \end{tabular}
     \caption{Distribution of the dataset with different number of documents in the input.}
    \label{tab:megadoc_distri}
\end{table}
\endgroup

\paragraph{Query Types} Although we explicitly pick the examples from \nq which \emph{only} have a long answer, we find many descriptive queries for factoid-like questions. For example, \textit{``Where is silver found and in what form (compound)''}, \textit{``Where was Moses when he saw a burning bush''}, \textit{``When is a jury used in civil cases''}. The most interesting examples are the \emph{why} queries since they have a descriptive summary in the long-form answer. For example, \textit{``Why does Friedman think the world is flat''}, \textit{``Why do plants drip water from their leaves''}.

\begin{figure}
  \centering
    \includegraphics[width=0.48\textwidth, viewport=50 35 600 270, clip=true]{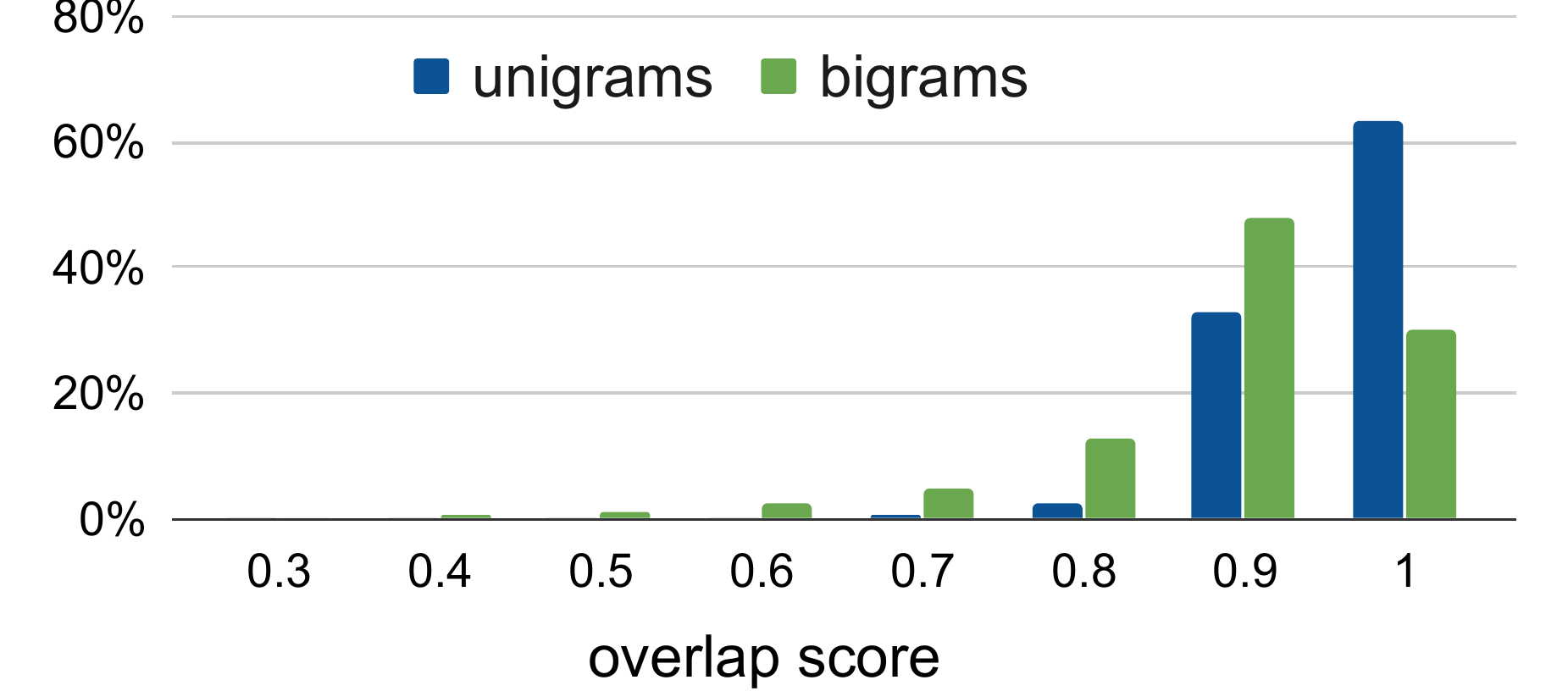}
    \caption{Overlap between the summary and input documents. 30\% of examples have bigram overlap score greater than 0.9 indicating novel bigrams in summary for over 70\% of cases.}
\label{fig:uni_bigram_overlap}
\end{figure}

\paragraph{Document and Summary Lexical Overlaps} 
Our approach relies on sentence-level matching to retrieve documents, but that does not guarantee a high recall of the n-grams in the summary. To sanity check that the set of no more than 7 documents retrieved this way has a high lexical overlap with the summary, we used the BLEU precision score. Note that a perfect BLEU precision score in this case implies that every n-gram in the summary can be mapped to a distinct n-gram in the source. Figure \ref{fig:uni_bigram_overlap} shows the histogram of this overlap measure.
\eat{Since our approach relies on the overlap between the summary and the source documents, it is crucial to know the lexical overlap between the two. Figure \ref{fig:uni_bigram_overlap} shows the histogram of the overlap measure. We use the modified BLEU score precision by treating the summary as the candidate and the \cc documents as the references. This metric has the benefit that it does not reward redundancy in the source documents and is an effective measure of collective coverage of the summary. \SC{confirm/rephrase}}

\paragraph{Comparing to other datasets} Table \ref{tab:compare_ds} compares our dataset instance with other commonly used datasets for summarization. CNN/DM is an abstractive SDS dataset \cite{nallapati2016abstractive}. Multi-News is the first large-scale MDS dataset \cite{fabbri2019multi}. While Multi-News has more summarization examples, our dataset includes query contexts and covers more documents, sentences and tokens per summary. Also the number \qmds examples can increase with looser restrictions on $\theta$ and $K$. Furthermore, our approach generalizes to more \emph{MDS} examples if we removed the query context and operated on any Wikipedia paragraph spans. Recently, \citet{nema-etal-2017-diversity} introduced a qMDS dataset built from Debatepedia\footnote{\url{http://www.debatepedia.org}}. Their input documents are relatively short (75 words/doc). \qmds includes much longer input documents that can be more challenging for qMDS models.

% !TEX root = main.tex
\section{Quality Assessment}
\label{sEval}
In this section, we carefully assess the quality of the automatically generated qMDS examples along several axes: correctness of matched documents, fluency of machine edited extractive summaries, and overall example quality. All our human evaluation tasks are based on human rater pools consisting of fluent English speakers. The annotation tasks are \emph{discriminative} in nature (e.g., judging semantic match), which are cheaper to source and easier to validate through replication than generative annotation tasks (e.g., open-ended text generation). We also provide a few qMDS examples for illustration.

\begin{figure}
  \centering
    \includegraphics[width=0.45\textwidth, viewport=50 45 690 270, clip=true]{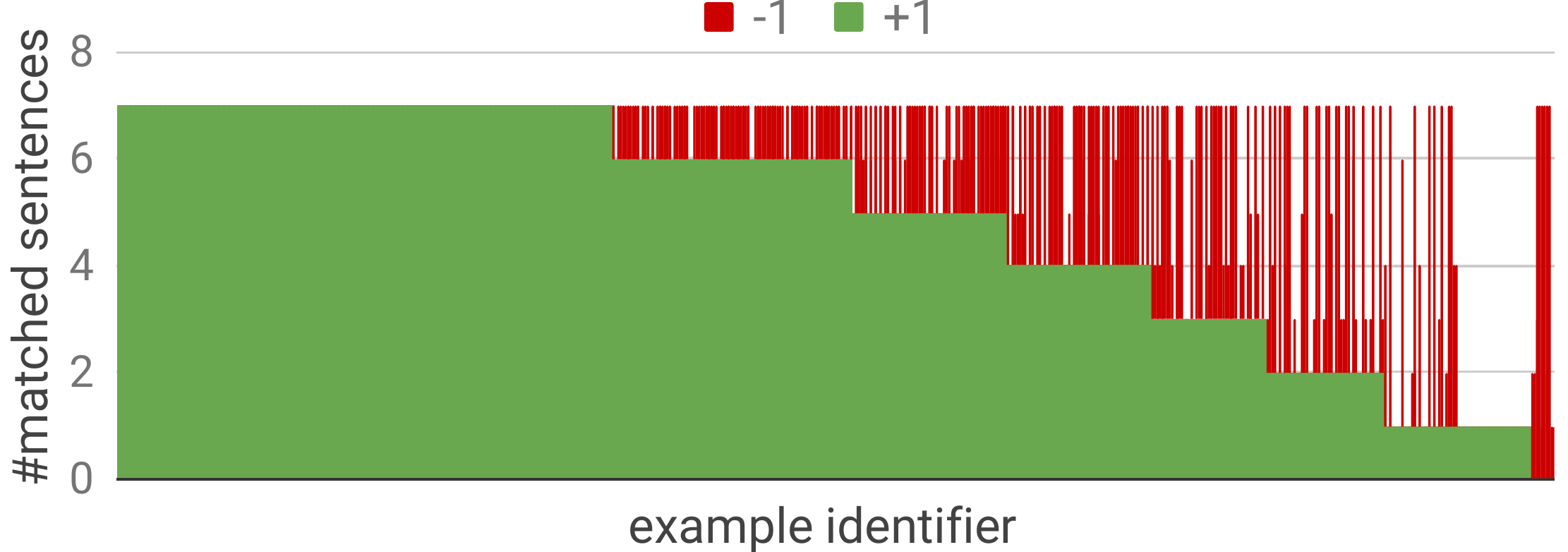}
  \caption{Majority decisions for sentence-to-document relevance task across examples rated where 1 sentence was sampled from each 7 matched \cc document. Examples are sorted by count of \texttt{+1} majority decision. Less relevant examples contain more raters abstaining from making a definitive decision.}
\label{fig:ratertask1_details}
\end{figure}

\subsection{Correctness}
We first evaluate the factual and semantic correctness of the \cc documents that were matched with the \nq long answers. We focus on the abstractive setup as we will demonstrate later how the derived extractive version is qualitatively similar.

For this annotation task, we presented raters with a Wikipedia paragraph (corresponding to the long-form answer) and a matched sentence (one from each of the top-7 \cc documents). They were asked to rate ``\texttt{+1}'' if the \cc sentence matched some of the content of the Wikipedia paragraph. Raters were instructed not to rely on external knowledge in the rating process. Numerical facts were subjectively evaluated, e.g., \textit{4B years} is close to \textit{4.5B years}, but \textit{3 electrons} and \textit{4 electrons} is not. 

\begin{figure}
  \centering
    \includegraphics[width=0.45\textwidth, viewport=30 40 610 315, clip=true]{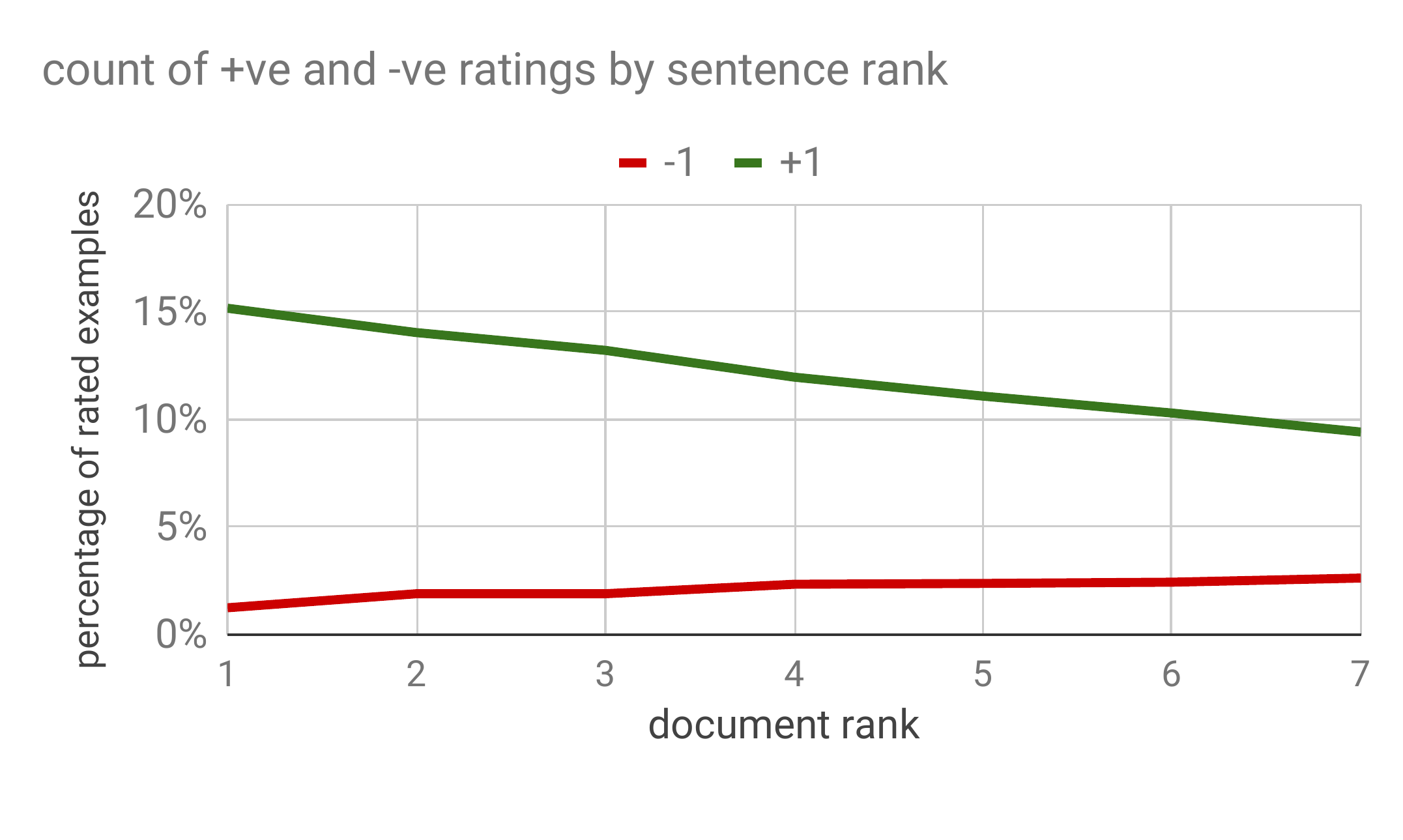}
  \caption{Higher ranked documents contain more sentences marked \texttt{+1} by majority. This is expected as high scoring documents should correlate with higher semantic relevance to the long answer.}
\label{fig:ratertask1_correl}
\end{figure}

We rated a sample of 5,215 examples corresponding to 856 queries. Each example rating is replicated 3 times across different raters to account for subjectivity. Raters were allowed to abstain if they cannot make a decision. We found that 85.18\% of the examples are marked relevant by majority as illustrated in Figure \ref{fig:ratertask1_details}. Sentences from top ranked documents (per document match scores) contains many more sentences annotated with \texttt{+1} majority decision as shown in Figure~\ref{fig:ratertask1_correl}.

\subsection{Fluency}
The extractive dataset is created by replacing sentences from the \cc doc with the matched sentence  in Wikipedia long answer $a$. This, however, may distort the overall flow of the original \cc passage. This evaluation task ensures that the fluency is not harmful.

First, we designed a human evaluation where the raters were presented with the original and the edited \cc document passages including the replaced sentence. A \texttt{+1} marks the replaced sentence \emph{does not} appear out of place. We rated 500 examples with rater replication of 3. In 96.20\% examples, these were rated positive.

Second, we measured the perplexity of the paragraphs that with replaced sentences using a language model \footnote{\url{https://tfhub.dev/google/wiki40b-lm-en/1}}. The mean perplexity increased slightly from 80 to 82 after replacement. This small increase is expected since a foreign sentence was inserted, but the difference is small enough proving that the fluency is preserved.

\subsection{Overall quality}
We now turn to evaluating the overall quality of a random sample of 55 qMDS example triplets $(q, R, a)$ along three dimensions --- referential clarity, focus and the coherence of the summary --- adapted from  DUC2007 task\footnote{\url{https://duc.nist.gov/duc2007/quality-questions.txt}}. Since the summary $a$ is  a Wikipedia passage, grammatical correctness and redundancy dimensions need not be evaluated.

\begin{table}
    \centering
    \begin{tabular}{r|ccccc|c}
            & 5 & 4 & 3 & 2 & 1 & NA \\
    \hline
    Clarity & 78 & 5 & 2 & 4 & 2 & 9 \\
    Focus & 67 & 13 & 4 & 0 & 0 & 16 \\
    Coherence & 71 & 11 & 0 & 4 & 2 & 13 \\
    \end{tabular}
     \caption{Distribution of clarity, focus and coherence ratings of abstractive qMDS examples on a 5-point scale. The scale runs from 5 (very good) to 1 (very poor), with NA designating inter-rater disagreements.}
    \label{tab:overall_ratings}
\end{table}

Each triplet was rated by 3 raters. The raters were also instructed to consider the query $q$ when evaluating the focus of the summary $a$ rather than just a generic summary that can be generated from the set of input documents $R$. Ratings were on a 5-point scale --- 5 being very good and 1 being very poor. The results are summarized in Table~\ref{tab:overall_ratings} showing that the majority of ratings fall under good (4) and very good (5).

\subsection{Examples}
Finally, we also illustrate two specific challenging aspects of the qMDS dataset. The example in Table \ref{tab:query2} demonstrates how a summary can cover multiple facets of a single query that can be sourced from multiple input documents. The example in Table \ref{tab:query3} shows how the query context may require summarization models to attend to specific portions of the source documents.

\begin{table}
\small
    \centering
    \begin{tabular}{p{0.47\textwidth}}
    \textbf{Query:} \highlightI{what} is a dream and \highlightII{why} do we dream \\
    \hline
    \textbf{Source1:} \highlightI{Dreams are successions of images, ideas, emotions and sensations occurring involuntarily in the mind during certain stages of sleep.} The content and purpose of dreams are not yet understood, though they have been a topic of speculation and interest throughout recorded history. \color{blue}{The scientific study of dreams is known as oneirology}... \\
    \textbf{Source2:} A dream is a succession of images ideas emotions and sensations that usually occur involuntarily in the mind during certain stages of sleep. \highlightII{The content and purpose of dreams are not fully understood although they have been a topic of scientific philosophical and religious interest throughout recorded history}. ...  \\
    \hline
    \textbf{Summary:} \highlightI{A dream is a succession of images, ideas, emotions, and sensations that usually occur involuntarily in the mind during certain stages of sleep.} \highlightII{The content and purpose of dreams are not fully understood, though they have been a topic of scientific speculation, as well as a subject of philosophical and religious interest, throughout recorded history.} Dream interpretation is the attempt at drawing meaning from dreams and searching for an underlying message. \highlightI{The scientific study of dreams is called oneirology}. \\
    \end{tabular}
    \caption{Multi-faceted query with summary generated using content merged from multiple sources.}
    \label{tab:query2}
\end{table}

% !TEX root = main.tex
\section{Experiments}
\label{sExperiments}
Our experiments are based on running popular summarization models on both abstractive and extractive versions of our qMDS dataset. These baseline summarization experiments are categorized into two types: (i) a \emph{query-agnostic} setup where the query $q$ is ignored and the models map 
 source documents $R$ to long answer $a$ as in standard SDS/MDS; and (ii) a \emph{query-based} setup where the source document set $R$ is conditionally filtered by the input query $q$ followed by SDS/MDS approaches.

\subsection{Abstractive Summarization}
\paragraph{Hi-MAP} \citet{fabbri2019multi} define a hierarchical abstractive MDS model that combines a pointer-generator network \cite{see2017get} with Maximal Marginal Relevance (MMR) \cite{carbonell1998use} scores to rank sentences based on relevancy and redundancy. We used 128-d word vectors in a single layer 512-d RNN that was trained up to 10K steps with an initial learning rate of 0.15.

\paragraph{PEGASUS} \citet{Zhang2019PEGASUSPW} propose pre-training Transformer-based MDS models with massive text corpora. Pre-training involves generating masked sentences, similar to an extractive summary. We fine-tune PEGASUS with an initial learning rate of 0.01 for 100K steps and evaluated it on our test set, with the caveat that some test \cc documents were part of the pre-trained model (albeit for a different objective).

\begin{table}
\small
    \centering
    \begin{tabular}{p{0.47\textwidth}}
    \textbf{Query:} characteristics of schoenberg’s atonal music \\
    \hline
    \textbf{Source1:} \textit{\hiddenHighlight{Paragraph\#1 is about the Musician. Paragraph\#2 is about the musicians approach.}} \highlightI{Schoenberg was widely known early in his career for his success in simultaneously extending the traditionally opposed German Romantic styles of both Brahms and Wagner.} Later, his name would come to personify pioneering innovations in atonality... \\
    \textbf{\emph{other sources...}} \\
    \hline
    \textbf{Summary:} \highlightI{Schoenberg was known early in his career for simultaneously extending the traditionally opposed German Romantic styles of Brahms and Wagner}. Later, his name would come to personify innovations in atonality (although Schoenberg himself detested that term) that would become the most polemical feature of 20th-century art music. In the 1920s, Schoenberg developed the twelve-tone technique, an influential compositional method of manipulating an ordered series of all twelve notes in the chromatic scale. He also coined the term developing variation and was the first modern composer to embrace ways of developing motifs without resorting to the dominance of a centralized melodic idea	 \\
    \end{tabular}
    \caption{Retrieving salient information related to the query in areas deep inside a source document and not necessarily the beginning of it.}
    \label{tab:query3}
\end{table}

\subsection{Extractive Summarization}
\paragraph{NeuSum} \citet{neusumm-zhou2018} rank sentences using scores derived from a hierarchical encoder, with top ranked sentences forming the extractive summary. While it was designed for SDS, the hierarchical document representation is well suited for adapting to the MDS setting in future work.
The model used 50-d GloVe word vectors that was trained with a learning rate of 0.001 and a batch size of 32 for 50 epochs. The output was set to 4 sentences to match the long answer summary statistics. Finally, the input sequence length was 500 sentences to capture the larger size of the multi-doc input.

\paragraph{TextRank} This is an unsupervised sentence similarity based summarization model based on weighted-graphs defined over sentences in a document~\cite{textrank2004} that is often used as a baseline for extractive summarization.

\subsection{Incorporating Query in SDS/MDS}
As our dataset explicitly designed for qMDS, we modified the standard SDS/MDS setup by pre-filtering sentences from the source documents $R$ that are relevant to query $q$ (based on BLEU scores) as input the models. To retain source document fluency, fragments are defined at the paragraph level. Table \ref{tab:absBaselines} and Table \ref{tab:extBaselines} show the results with and without this variation. The filter acts as a crude attention mechanism that weeds out irrelevant content from the inputs showing improvements in all the approaches, except for PEGASUS. We believe this drop may be attributed to the sentence masking done in pre-training PEGASUS which relies on undisrupted sentence orders. 

\subsection{Human Evaluation}
In addition to automatic evaluation, we also collected human judgements for summarization outputs of one specific abstractive MDS model (Hi-MAP) to understand the headroom available in qMDS on this dataset. We follow the question-answering approach in \citet{clarke-lapata-2010}.

We created 32 questions from 17 randomly sampled summaries. Participants are asked to answer those questions after reading the \emph{generated} summary by Hi-MAP.  Their answers are scored: 1 (fully correct answer), 0.5 (partially correct answer), and 0 (incorrect answer). Note that the ground-truth answers are the answers to the ground-truth summaries. The more the participants can answer correctly from the generated summaries, the better the summarization system. We then compute the averaged scores. 5 were given questions and the ground-truth summaries and the rest 5 were given the questions and the generated summaries by Hi-MAP. For ground-truth summaries, the average positive responses is 30.6 out of 32. For Hi-MAP summary, this is 13.8. This is significantly lower than the score for  ground-truth showing a fairly wide headroom for improvement.

\begin{table}
    \centering
    \begin{tabular}{l | ccc}
    Method &  R-1 & R-2 & R-L \\
    \hline
    \multicolumn{4}{c}{Query-agnostic setting} \\
    \hline
    Hi-MAP & 28.34 & 13.12 & 25.15 \\
    PEGASUS & 27.08 & 12.51 & 22.28  \\
    \hline
\multicolumn{4}{c}{Query-based setting } \\
    \hline
    Hi-MAP & 30.34 & 14.82 & 26.86 \\
    PEGASUS & 24.61 & 9.12 & 19.61 \\
    \end{tabular}
    \caption{Baselines on abstractive dataset on test split }
    \label{tab:absBaselines}
\end{table}

\begin{table}
    \centering
    \begin{tabular}{l | ccc}
    Method &  R-1 & R-2 & R-L \\
    \hline
    \multicolumn{4}{c}{Query-agnostic setting} \\
    \hline
    NeuSum  & 62.61 & 54.45  & 61.99 \\
    TextRank & 24.4 & 15.56 & 31.6 \\
    \hline
    \multicolumn{4}{c}{Query-based setting} \\
    \hline
    NeuSum & 63.09 & 55.13  & 62.39 \\
    TextRank & 25.72 & 17.4 & 34.3 \\
    \end{tabular}
     \caption{Baselines on extractive dataset on test split.}
     \label{tab:extBaselines}
\end{table}

% !TEX root = main.tex
\section{Related Work}
\label{sRelated}

Query-based summarization can be both extractive~\cite{dang2005duc,daume-iii-marcu-2006-bayesian,Schilder-fastsum,Otterbacher2009,wang-etal-2013-sentence,litvak-2017-query,wang-etal-2019-self} or abstractive~\cite{nema-etal-2017-diversity,baumel-arxiv_1801.07704,abstractive-nn-qbs,Ishigaki1-2020-ecir}. Earlier studies were often extractive and relied on manually selected and curated datasets such as DUC2005 and DUC2006 \cite{dang2005duc}. However, neural abstractive models often demand large amounts of labeled data, which are hard to obtain for summarization and other tasks with similar demands on manual annotation efforts. Recent studies show a two-step process of using extractive summarization followed by generation for abstractive summaries\cite{fabbri2019multi} as well as for query-based abstractive summaries~\cite{egonmwan-etal-2019-cross}. 

While our work is motivated by the use case of generating longer summaries to answer complex questions, there are related work on creating QA datasets for short answers:  using news articles from CNN/DM \cite{hermann15nips},  HotpotQA \cite{yang-etal-2018-hotpotqa}, TriviaQA~\cite{joshi-etal-2017-triviaqa}, SearchQA~\cite{dunn2017searchqa}, online debates with summarizing arguments and debating topics as queries \cite{nema-etal-2017-diversity}, and community question answering websites \cite{deng2020aaai20}. Some of them involve extracting text spans of words as answers from multiple text passages. However, our work focuses on longer answers. 

Large-scale datasets for regular MDS over long documents with target long summaries have also started to appear \cite{liu2018generating,fabbri2019multi,koupaee2018wikihow}. Besides extracting contents for IR applications, our efforts differs from them in terms of heterogeneity in documents and lengths of summaries. The MS Marco dataset is close to our work in spirit~\cite{bajaj2018ms}. The dataset contains 1M question-answer-context triplets where the answers are human created using the top-10 passages returned from Bing's search queries. We use Wikipedia passages as summaries thus avoid additional human efforts. Examining the statistics of the dataset, our dataset also has longer input sources and answers.

% !TEX root = main.tex
\section{Conclusion}
\label{sConclusion}
We have presented \qmds, a scalable methodology for constructing new qMDS datasets, along with in-depth analyses and baseline experiments to demonstrate properties of one such dataset instance. Many parts of the approach are configurable providing researchers a rich sandbox for evaluating summarization models under different task conditions. Our methodology greatly reduces the cost of data collection by converting a predominantly generative human annotation task (e.g., reading documents and writing succinct summaries) to a discriminative human annotation task (e.g., deciding on sentence-document relevance). While our present work do not propose new methods for query-based summarization, we ran baseline experiments on one specific instance of the \qmds dataset using a few popular neural approaches re-adapted with query conditioning. Our experiments demonstrates that there is still much headroom for existing state-of-the-art models and we hope \qmds will spur further advancements query focused multi-document summarization algorithms.

\bibliography{qbs-ds}
\bibliographystyle{acl_natbib}

\end{document}